\documentclass{article}

\usepackage{microtype}
\usepackage{graphicx}
\usepackage{subfigure}
\usepackage{booktabs} 
\usepackage{multirow}
\usepackage{colortbl}
\usepackage{makecell}
\usepackage{rotating}
\usepackage{enumitem}

\usepackage{hyperref}

\usepackage[preprint]{icml2026}

\usepackage{amsmath}
\usepackage{amssymb}
\usepackage{mathtools}
\usepackage{amsthm}

\usepackage[capitalize,noabbrev]{cleveref}

\theoremstyle{plain}
\newtheorem{theorem}{Theorem}[section]

\theoremstyle{definition}
\newtheorem{definition}[theorem]{Definition}
\newtheorem{assumption}[theorem]{Assumption}
\theoremstyle{remark}

\usepackage[textsize=tiny]{todonotes}

\usepackage[acronym]{glossaries}
\glsdisablehyper
\makeglossaries
\newacronym{rlhf}{RLHF}{Reinforcement Learning from Human Feedback}
\newacronym{dpo}{DPO}{Direct Preference Optimization}
\newacronym{sft}{SFT}{supervised fine-tuning}
\newacronym{hrl}{HRL}{high-resource language}
\newacronym{lrl}{LRL}{low-resource language}
\newacronym{llm}{LLM}{Large Language Model}

\icmltitlerunning{Multilingual Safety Alignment Via Sparse Weight Editing}

\begin{document}

\twocolumn[
\icmltitle{Multilingual Safety Alignment Via Sparse Weight Editing}



\icmlsetsymbol{equal}{*}

\begin{icmlauthorlist}
\icmlauthor{Jiaming Liang}{yyy}
\icmlauthor{Zhaoxin Wang}{yyy}
\icmlauthor{Handing Wang}{yyy}

\end{icmlauthorlist}

\icmlaffiliation{yyy}{School of Artificial Intelligence, Xidian University}

\icmlcorrespondingauthor{Handing Wang}{hdwang@xidian.edu.cn}

\icmlkeywords{Multilingual, Large Language Model, Safety Alignment}

\vskip 0.3in
]



\printAffiliationsAndNotice{\icmlEqualContribution} 

\begin{abstract}
\glspl{llm} exhibit significant safety disparities across languages, with \glspl{lrl} often bypassing safety guardrails established for \glspl{hrl} like English. Existing solutions, such as multilingual \gls{sft} or \gls{rlhf}, are computationally expensive and dependent on scarce multilingual safety data. In this work, we propose a novel, training-free alignment framework based on \textit{Sparse Weight Editing}. Identifying that safety capabilities are localized within a sparse set of "safety neurons", we formulate the cross-lingual alignment problem as a constrained linear transformation. We derive a closed-form solution to optimally map the harmful representations of LRLs to the robust safety subspaces of HRLs, while preserving general utility via a null-space projection constraint. Extensive experiments across 8 languages and multiple model families (Llama-3, Qwen-2.5) demonstrate that our method substantially reduces Attack Success Rate (ASR) in LRLs with negligible impact on general reasoning capabilities, all achieved with a single, data-efficient calculation.
\end{abstract}

\section{Introduction}
The rapid advancement of large language models (\glspl{llm}) has enabled impactful applications across domains\cite{achiam2023gpt, yang2025qwen3}. However, when deployed in open and interactive environment, \glspl{llm} are exposed to diverse threats, raising safety concerns~\cite{chander2025toward}. For example, adversarial attacks~\cite{szegedy2013intriguing,goodfellow2014explaining,wang2024preventing} can undermine reliability, while backdoor attacks can trigger malicious behaviors via data poisoning~\cite{gu2019badnets}. Moreover, adversaries can exploit jailbreak attacks~\cite{yi2024jailbreak,wang2025implicit} to elicit harmful outputs.

To mitigate these risks, researchers have developed safety alignment techniques~\cite{leike2018scalable,kenton2021alignment,ji2023ai}, including reinforcement learning from human feedback (\gls{rlhf})~\cite{rlhf,rlhf2} and preference optimization methods~\cite{dpo,shao2024deepseekmath}, to align model behavior toward human values and social norms. Despite their effectiveness, these methods are data-intensive, requiring large-scale, carefully curated preference datasets, which are expensive and time-consuming to collect. This challenge is particularly acute in multilingual settings, as such datasets are abundant for high-resource languages (\glspl{hrl}) like English but scarce for many low-resource languages (\glspl{lrl}), leading to substantial cross-lingual disparities in safety. The same \gls{llm} is often well-aligned in English but considerably less safe in \glspl{lrl}.

To bridge the gap of multilingual safety, the recent work~\cite{bu2025alignx,mpo,adamergex} leverages multilingual corpora to improve safety in \glspl{lrl}, often by relying on supervised fine-tuning (\gls{sft})~\cite{sft}. However, these methods  depend on costly, high-quality safety datasets in multiple languages. Some work~\cite{linguistic} transfers safety capabilities from \glspl{hrl} to \glspl{lrl} through intermediate languages bridge, but generally assumes strong translation performance. In practice, translation errors can propagate to downstream reasoning and generation, and translation-based pipelines introduce additional inference overhead.

Recent studies \cite{linguistic} have identified the existence of "linguistic overlap neurons", specific neurons that are activated by both \glspl{hrl} and \glspl{lrl} and play a pivotal role in encoding core model capabilities, including safety mechanisms~\cite{zhao2025understanding}. This observation introduces a critical question: 
\begin{center}
    \textbf{Can we transfer the safety representations of \glspl{hrl} to \glspl{lrl} without retraining?}
\end{center}

In this work, we propose a multilingual alignment framework that transfers safety capabilities learned from \glspl{hrl} (e.g., English) to \glspl{lrl}. Concretely, we parameterize the cross-lingual adjustment as a low-rank transformation in representation space, and solve a transformation matrix that maps the feature representations of harmful queries in \glspl{lrl} to the well-aligned, safe activation patterns of \glspl{hrl}. To ensure this modification does not compromise the model's utility, we introduce a {null-space projection constraint} derived from harmless data. This constraint ensures that our intervention is orthogonal to the directions encoding general capabilities, thereby modifying the safety-critical feature subspace, while minimizing effects on general capabilities.

Distinguishing our approach from prior work, we derive a closed-form solution to this optimization problem. This allows us to compute the optimal alignment parameters analytically using only a few anchor samples, eliminating the need for gradient-based training. Our contributions are summarized as follows: 

    \begin{itemize}[leftmargin=1.5em]
    \item \textbf{Representation-Level Safety Transfer.} We introduce a representation alignment method for multilingual safety that maps the well-aligned and safe activation patterns from \glspl{hrl} to \glspl{lrl} tasks, thereby providing safety improvements across languages. 
    \item \textbf{Training-Free Efficiency.} We formulate cross-lingual safety alignment as a regularized low-rank update problem and derive a closed-form solution. Our method requires only a small number of harmful and harmless anchor samples to compute the modification matrix, avoiding iterative gradient-based optimization.
    \item \textbf{Interpretable and Plug-and-Play Intervention.} Our framework provides an interpretable view of transferable safety-related representation components across languages. Moreover, it acts as a lightweight, plug-and-play intervention that can be integrated into different model architectures without disrupting parameters.
    \end{itemize}

\section{Related Works}

\subsection{Jailbreak Attacks}

Despite their remarkable capabilities, \glspl{llm} remain vulnerable to adversarial exploitation. Attackers can craft \emph{jailbreak} inputs---carefully engineered prompts designed to circumvent safety alignment and elicit harmful behaviors. Existing black-box jailbreak strategies primarily exploit the instruction-following nature of \glspl{llm} via sophisticated input manipulation and automated prompt search~\cite{yi2024jailbreak}. Static approaches often leverage models' pattern-completion tendency by embedding malicious requests within benign-looking templates~\cite{li2023deepinception,yao2024fuzzllm,anil2024many,wei2023jailbreak}. Such inputs may evade superficial safety filters while remaining interpretable to the underlying model. More advanced paradigms shift toward automated red-teaming, framing jailbreaking as an optimization problem. Using auxiliary \glspl{llm} together with heuristics such as genetic algorithms or gradient-free optimization methods, these methods iteratively refine adversarial prompts to maximize attack success rate~\cite{liu2024autodan,mehrotra2024tree,chao2025jailbreaking}.

\subsection{Multilingual Safety Enhancement}
\paragraph{Training time.}
Recent work extends safety alignment to multilingual settings by constructing cross-lingual safety datasets or leveraging \glspl{hrl} signals as supervision. For example, AlignX~\cite{bu2025alignx} proposes a two-stage framework that first aligns multilingual representations and then fine-tunes the model with multilingual instructions to reduce the performance gap between \glspl{hrl} and \glspl{lrl}. Similarly, MPO~\cite{mpo} introduces a multilingual reward-gap optimization objective that minimizes discrepancies between reward distributions in \glspl{hrl} (e.g., English) and \glspl{lrl}, thereby facilitating cross-lingual safety transfer. AdaMergeX~\cite{adamergex} explores cross-lingual transfer via adaptive adapter merging, aiming to decouple task competence from language competence. 
\paragraph{Inference time.}
To circumvent the high costs of retraining, researchers have investigated inference-time interventions and parameter-efficient strategies. For example, RESTA \cite{resta} employs the task arithmetic to recover safety by adding a pre-computed safety vector—derived from the difference between an aligned and a deliberately unaligned model—to task-specific \glspl{llm}. However, this approach has several intrinsic drawbacks. First, the initial extraction of the safety vector necessitates a risky unalignment process and relies heavily on the coverage of the harmful datasets used. Second, the linear arithmetic operation on model weights lacks fine-grained control over specific linguistic neurons, often leading to a suboptimal trade-off between safety enforcement and the preservation of general capabilities.
\paragraph{Translation-based mehtods.}
Given the dominance of English-centric safety alignment, a widely used strategy is the Translate-Test pipeline~\cite{ponti2021modelling,artetxe2023revisiting,etxaniz2024multilingual}, which translates \glspl{lrl} inputs into English for safety processing. Extensions such as BridgeX-ICL~\cite{linguistic} further improve cross-lingual transfer by routing through bridge languages and exploiting linguistic overlap neurons. Despite their simplicity, translation-based methods face a fundamental semantic bottleneck that safety-critical intent may be altered or lost during translation, making safety enforcement in the original language less reliable. 

\subsection{Neuron Identification}
Research on neuron-level interpretability has shifted from characterizing general model capabilities \cite{dai2022knowledge, wang2022finding} to identifying "Safety Neurons" critical for alignment. Current approaches typically locate these neurons through inference-time activation contrasting \cite{chen2024finding}, linear probing classifiers \cite{wu2025neurostrike}, or ablation-based importance scoring \cite{zhao2025understanding}. While findings on specific layer distribution vary, ranging from Feed-Forward Networks \cite{chen2024finding} to Self-Attention layers \cite{zhao2025understanding}. These studies collectively establish that safety mechanisms are highly sparse, relying on less than 1\% of total parameters to suppress harmful content effectively \cite{zhao2025understanding, wu2025neurostrike,wang2026safeneuronneuronlevelsafetyalignment}.

\section{Empirical findings}
\label{sec:empfindings}

\subsection{Definition of Safety Neurons}
Prior research \cite{chen2024finding,marks2024sparse,dunefsky2024transcoders} in mechanistic interpretability suggests that the high-level capabilities of LLMs are often localized within specific, sparse sub-structures of the network. Building on this, we revisit "can we transfer the safety representations of \glspl{hrl} to \glspl{lrl} without retraining?"

\begin{assumption}[Sparse Safety Localization] 
\label{assum:sparse} Motivated by~\cite{wu2025neurostrike,wang2026safeneuronneuronlevelsafetyalignment}, we assume that safety-related behavior in \glspl{llm} can be effectively influenced through a sparse subset of neurons within the Multilayer Perceptron (MLP) layers. These neurons, denoted as \textit{Safety Neurons}, exhibit significant activation divergence when processing harmful versus harmless inputs. \end{assumption}

To identify these neurons, we employ a dual-metric procedure for MLP activations (\texttt{up\_proj} and \texttt{gate\_proj}), following prior neuron-identification practice (including a concurrent submission by the authors~\cite{wang2026safeneuronneuronlevelsafetyalignment}). 
Specifically, by contrasting activations under harmful and harmless queries, we select units that exhibit both a large absolute activation gap and strong statistical separability. 
Detailed formulations and extraction hyperparameters are provided in Appendix~\ref{app:neuron_extraction}.

\subsection{Activation Steering for Cross-Lingual Safety}
\label{sec:steering_experiment}
To verify whether the identified English safety neurons $\mathcal{S}_{eng}$ play a functional role in multilingual safety, we conduct an activation steering experiment. Our intuition is that if English acts as a dominant semantic anchor during training, strengthening English safety-related activations may improve safety behavior in other languages.

Specifically, during the forward pass of the model, we intervene on the activations of the identified safety neurons. For every neuron $j \in \mathcal{S}_{eng}$ at layer $l$, we scale its output activation $x_{j}^{(l)}$ by a coefficient $\alpha > 1$:
\begin{equation} 
	\tilde{x}_{j}^{(l)} = \alpha \cdot x_{j}^{(l)} \quad \forall j \in \mathcal{S}_{eng},
\end{equation}
where $\tilde{x}_{j}^{(l)}$ is the intervened activation. We evaluate the model's attack success rate (ASR) on multilingual jailbreak prompts under varying scaling factors $\alpha$.

\begin{figure}[ht] 
\centerline{\includegraphics[width=1\columnwidth]{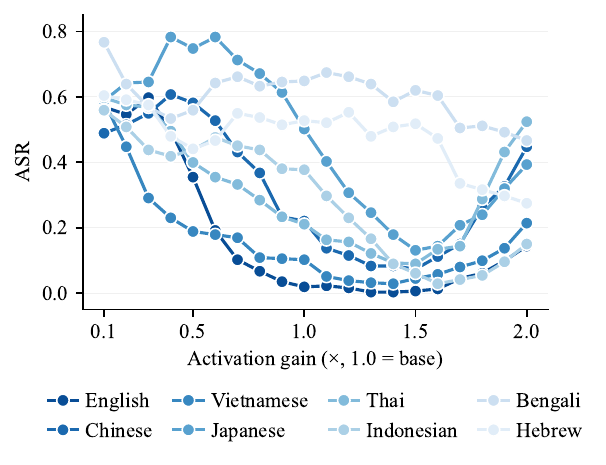}} 
\caption{\textbf{Impact of English Safety Neuron Amplification.} Scaling the activations of English safety neurons leads to a consistent decrease in harmful response rates across multiple languages, validating the cross-lingual influence of these neurons.} 
\label{fig:steering_results} 
\end{figure}

As illustrated in Figure \ref{fig:steering_results}, simply amplifying English safety neurons significantly improves safety across various languages. This confirms that English safety neurons act as a universal safety neurons to some extent, leveraging the model's cross-lingual alignment.

\vspace{-0.2cm}
\subsection{Representation Transfer}
\label{sec:transfer}

\begin{figure}[h]
\vskip 0.2in
\centerline{\includegraphics[width=0.9\columnwidth]{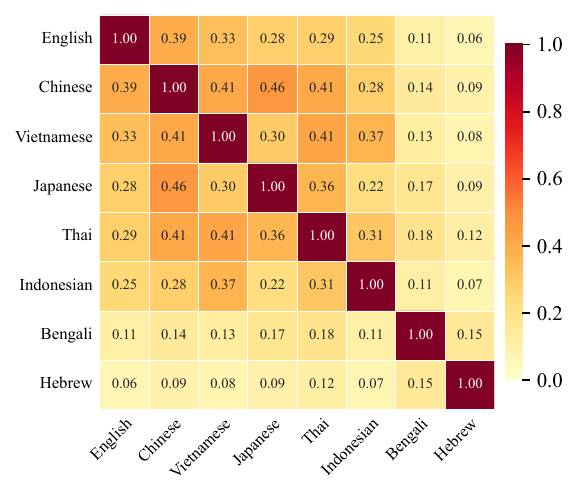}} 
\caption{Pairwise safety-neuron set overlap across languages. Higher values indicate greater overlap under this set-based measure. \glspl{hrl} tend to exhibit stronger overlap, whereas \glspl{lrl} show weaker overlap both with \glspl{hrl} and with each other.}
\label{fig:overlap_correlation} 
\end{figure}

While simple amplification can improve safety in some cases, we observe substantial variation in its effectiveness across languages. To further investigate the reason for this variability, we examine whether it is associated with \textbf{cross-lingual representation overlap}. We compute the \textbf{Jaccard similarity} (intersection over union) between the safety-neuron sets identified for each pair of languages. Formally, for languages $\ell_i$ and $\ell_j$, we define
\begin{equation}
  \mathrm{Jaccard}(\mathcal{S}_{\ell_i}, \mathcal{S}_{\ell_j})
=
\frac{|\mathcal{S}_{\ell_i} \cap \mathcal{S}_{\ell_j}|}{|\mathcal{S}_{\ell_i} \cup \mathcal{S}_{\ell_j}|},  
\end{equation}
where $\mathcal{S}_{\ell}$ denotes the safety-neuron index set extracted for language $\ell$. Figure~\ref{fig:overlap_correlation} visualizes these pairwise similarities as a heatmap.

The heatmap reveals a clear overlap pattern that high-resource languages exhibit consistently higher \emph{safety-neuron set} overlap, whereas low-resource languages show weaker overlap, both with high-resource languages and with one another. English has relatively high Jaccard similarity with several other languages, while many low-resource languages display more limited overlap and appear more isolated under this set-based similarity measure.

This observation explains the limitations of simple activation steering. When a target language already activates a safety-neuron subset aligned with the English-centric safety subspace, amplifying those neurons effectively suppresses harmful generation. However, for languages whose safety-relevant features are distributed over a distinct set of neurons, amplification primarily increases the magnitude of a mismatched activation pattern without correcting its direction. These findings highlight a fundamental geometric limitation of passive transfer mechanisms, that safety representations are not universally aligned across languages. As a result, effective multilingual safety alignment requires an \emph{active reorientation} of language-specific safety representations toward a shared, robust safety anchor, rather than relying on incidental neuron overlap.

\vskip -0.2in
\section{Method}

Motivated by the empirical findings in Section~\ref{sec:empfindings}, we propose \textsc{Sparse Weight Editing}, a training-free alignment framework for bridging the representation gap between \glspl{hrl} and \glspl{lrl}. Our key observation is that the English safety subspace provides a reliable alignment anchor (Section~\ref{sec:steering_experiment}), whereas many \glspl{lrl} exhibit directional misalignment that makes simple activation steering ineffective (Section~\ref{sec:transfer}). We therefore cast cross-lingual safety transfer as a constrained linear transformation problem: we compute a sparse perturbation $\Delta W$ that aligns harmful representations in \glspl{lrl} toward the safe activation patterns of \glspl{hrl}, while preserving general utility via a null-space constraint.

\subsection{Safety Neuron Identification}
To identify the safety-critical neurons for each language, we construct a multilingual probing dataset by translating standard harmful ($\mathcal{D}_{harm}$) and harmless ($\mathcal{D}_{safe}$) corpora into our target languages (Appendix~\ref{app:dataset}). These language-specific probes enable us to contrast activations under harmful versus harmless inputs and localize the sparse neuronal subset most associated with safety behaviors, which we subsequently target in our weight editing procedure.

\subsection{Cross-Lingual Safety Subspace Alignment}
\label{sec:motivation_bridge}

The empirical results in Section~\ref{sec:transfer} reveal a fundamental limitation of direct safety transfer: \emph{representation misalignment}. While HRLs such as English activate a distinct safety subspace, i.e., a characteristic activation pattern over safety neurons, LRL queries often induce feature representations that are orthogonal to or deviated from this subspace, likely due to insufficient safety supervision in the target language. As a result, simple activation amplification (Section~\ref{sec:transfer}) is ineffective for low-overlap languages. It increases the magnitude of an already misaligned representation without correcting its direction.

To address this, we explicitly \emph{reorient} harmful representations in LRLs toward the safety pattern of HRLs via a weight-space linear mapping. Concretely, we solve for a sparse perturbation $\Delta W_{\mathcal{S}}$ applied to the safety weight submatrix $W_{\mathcal{S}}$ such that the projected activations for LRL harmful inputs $X_{low}$ match the target safety activations $Y_{target}$ derived from aligned HRLs:
\begin{equation}
    \sigma\!\left(X_{low}\left(W_{\mathcal{S}} + \Delta W_{\mathcal{S}}\right)\right) \approx Y_{target}.
\label{eq:subspace_align}
\end{equation}
Here, $Y_{target}$ represents the desired safety activation pattern (e.g., the activations of English safety neurons under corresponding harmful queries). By minimizing the reconstruction error between the transformed LRL activations and $Y_{target}$, we enable cross-lingual safety transfer without retraining the full model.

\subsection{Weight-Editing Formulation}
\label{sec:problem_formulation}

We formulate cross-lingual safety transfer as a lightweight weight-editing problem on a small, safety-relevant subspace. The key idea is to (i) restrict the update to the identified safety neurons for parameter efficiency, (ii) align harmful LRL representations toward an English-derived safety activation target, and (iii) preserve benign utility via a null-space regularization. Finally, we impose a low-rank structure to improve robustness in the few-shot regime.

\subsubsection{Subspace Selection}
To minimize interference with general capabilities, we restrict weight editing strictly to the identified safety neurons. Let $\mathcal{S}$ denote the index set of safety neurons at layer $l$, with $|\mathcal{S}| = m$. The original weight matrix is $\boldsymbol{W} \in \mathbb{R}^{d_{in} \times d_{out}}$. We define the safety weight submatrix $\boldsymbol{W}_{\mathcal{S}} \in \mathbb{R}^{d_{in} \times m}$ as the columns of $\boldsymbol{W}$ indexed by $\mathcal{S}$. Our goal is to learn a perturbation $\Delta \boldsymbol{W}_{\mathcal{S}} \in \mathbb{R}^{d_{in} \times m}$ applied only to these columns, while keeping the remaining weights $\boldsymbol{W}_{\setminus \mathcal{S}}$ frozen.

\subsubsection{Alignment Objective}
We aim to align the harmful representations of \glspl{lrl} with the safety activation patterns induced by English. Let $\boldsymbol{X}_{\text{low}} \in \mathbb{R}^{N_h \times d_{in}}$ denote the layer-$l$ input features extracted from harmful LRL queries, and let $\boldsymbol{Y}_{\text{target}} \in \mathbb{R}^{N_h \times m}$ denote the target activations of the safety neurons derived from the English context. We seek $\Delta \boldsymbol{W}_{\mathcal{S}}$ such that
\begin{equation}
\sigma\!\left(\boldsymbol{X}_{\text{low}}\left(\boldsymbol{W}_{\mathcal{S}} + \Delta \boldsymbol{W}_{\mathcal{S}}\right)\right) \approx \boldsymbol{Y}_{\text{target}}.
\label{eq:nonlinear_target}
\end{equation}

Directly optimizing Eq.~\ref{eq:nonlinear_target} is inconvenient due to the nonlinearity $\sigma(\cdot)$. Following the motivation in Section~\ref{sec:motivation_bridge}, we adopt a first-order approximation in the pre-activation space and minimize the residual of the linear term:
\begin{equation}
    \mathcal{L}_{\text{align}}=
    \left\|
        \boldsymbol{X}_{\text{low}} \Delta \boldsymbol{W}_{\mathcal{S}}-
        \left(\boldsymbol{Y}_{\text{target}} - \boldsymbol{X}_{\text{low}} \boldsymbol{W}_{\mathcal{S}}\right)
    \right\|_{F}^{2}.
\label{eq:align_loss}
\end{equation}

Define the \emph{Safety Gap} as
\begin{equation}
\boldsymbol{D}_{\mathcal{S}}
=
\boldsymbol{Y}_{\text{target}} - \boldsymbol{X}_{\text{low}} \boldsymbol{W}_{\mathcal{S}}
\in \mathbb{R}^{N_h \times m},
\label{eq:safety_gap}
\end{equation}
which captures the activation discrepancy that $\Delta \boldsymbol{W}_{\mathcal{S}}$ is expected to bridge.

\subsubsection{Utility Constraint and Regularization}
To preserve benign-task performance, we introduce a utility-preserving null-space regularization. Let $\boldsymbol{X}_{\text{safe}} \in \mathbb{R}^{N_s \times d_{in}}$ denote the layer-$l$ input features extracted from harmless queries. We encourage the perturbation $\Delta \boldsymbol{W}_{\mathcal{S}}$ to lie in the (right) null space of $\boldsymbol{X}_{\text{safe}}$, so that it induces minimal change on harmless features:
\begin{equation}
\mathcal{L}_{\text{utility}}
=
\left\|
\boldsymbol{X}_{\text{safe}} \Delta \boldsymbol{W}_{\mathcal{S}}
\right\|_{F}^{2}.
\label{eq:utility_loss}
\end{equation}

\subsubsection{Low-Rank Constraint}
Optimizing a dense, full-rank perturbation $\Delta \boldsymbol{W}_{\mathcal{S}}$ is undesirable in our few-shot regime, where the anchor set is small relative to the number of free parameters. Without additional structure, a full-rank update can overfit to noise and exhibit poor generalization. Moreover, consistent with the \emph{Sparse Safety Localization} assumption (Assumption~\ref{assum:sparse}), we expect the safety-relevant update to be concentrated in a low-dimensional subspace. We therefore impose a low-rank constraint $\operatorname{rank}(\Delta \boldsymbol{W}_{\mathcal{S}})\le r$, which encourages the update to modify only the principal directions.

Combining the alignment objective, the utility regularization, and weight decay, the final optimization problem is:
\begin{equation}
\label{eq:final_optimization}
\begin{aligned}
\min_{\Delta \boldsymbol{W}_{\mathcal{S}}}  &
\left\|\boldsymbol{X}_{\text{low}} \Delta \boldsymbol{W}_{\mathcal{S}} -\boldsymbol{D}_{\mathcal{S}} \right\|_{F}^{2}& + \gamma \left\|\boldsymbol{X}_{\text{safe}} \Delta \boldsymbol{W}_{\mathcal{S}} \right\|_{F}^{2} \\
& + \lambda \left\|\Delta \boldsymbol{W}_{\mathcal{S}}\right\|_{F}^{2} \\[0.2cm]
\text{s.t.} \quad & \operatorname{rank}(\Delta \boldsymbol{W}_{\mathcal{S}}) \leq r .
\end{aligned}
\end{equation}
Since $\Delta \boldsymbol{W}_{\mathcal{S}} \in \mathbb{R}^{d_{in}\times m}$ only edits the columns corresponding to safety neurons, the update is lightweight compared to modifying the full weight matrix.

\subsection{Closed-Form Solution}
\label{sec:closed_form}

A key advantage of Eq.~\ref{eq:final_optimization} is that it admits an analytic solution, avoiding iterative gradient-based optimization. In this section, we show that the rank constraint can be handled by reducing the objective to a standard low-rank approximation under a whitened metric, which yields an efficient single-pass solver.

\begin{theorem}[Low-Rank Safety Alignment]
\label{thm:closed_form}
Define
\begin{equation}
\boldsymbol{Q}
=
\boldsymbol{X}_{\text{low}}^{\top}\boldsymbol{X}_{\text{low}}
+
\gamma\,\boldsymbol{X}_{\text{safe}}^{\top}\boldsymbol{X}_{\text{safe}}
+
\lambda\,\boldsymbol{I}.
\label{eq:Q_def}
\end{equation}
For $\lambda>0$, $\boldsymbol{Q}$ is positive definite and admits a Cholesky factorization $\boldsymbol{Q}=\boldsymbol{R}^{\top}\boldsymbol{R}$.
Let
\begin{equation}
\boldsymbol{M}
=
\boldsymbol{Q}^{-1}\boldsymbol{X}_{\text{low}}^{\top}\boldsymbol{D}_{\mathcal{S}}
\label{eq:M_def}
\end{equation}
be the optimal solution of Eq.~\ref{eq:final_optimization} \emph{without} the rank constraint.
Then the optimal rank-$r$ perturbation $\Delta \boldsymbol{W}_{\mathcal{S}}^{*}$ for Eq.~\ref{eq:final_optimization} is
\begin{equation}
\Delta \boldsymbol{W}_{\mathcal{S}}^{*}
=
\boldsymbol{R}^{-1}\tilde{\boldsymbol{\Delta}}^{*},
\label{eq:closed_form_sol}
\end{equation}
where $\tilde{\boldsymbol{\Delta}}^{*}$ is the best rank-$r$ approximation of $\tilde{\boldsymbol{M}}=\boldsymbol{R}\boldsymbol{M}$ in Frobenius norm. Concretely, if $\tilde{\boldsymbol{M}}=\boldsymbol{U}\boldsymbol{\Sigma}\boldsymbol{V}^{\top}$ is the SVD of $\tilde{\boldsymbol{M}}$, then
\begin{equation}
\tilde{\boldsymbol{\Delta}}^{*}
=
\boldsymbol{U}\boldsymbol{\Sigma}_{r}\boldsymbol{V}^{\top},
\label{eq:rank_r_svd}
\end{equation}
with $\boldsymbol{\Sigma}_{r}$ keeping only the top-$r$ singular values (and setting the rest to zero).
\end{theorem}

See Appendix~\ref{app:proof} for the derivation based on the Eckart--Young--Mirsky theorem.

\begin{algorithm}[htb]
\caption{Closed-Form Solver for Sparse Weight Editing}
\label{alg:closed_form_compact}
\begin{algorithmic}[1]
\STATE {\bfseries Input:} $\boldsymbol{X}_{\text{low}}, \boldsymbol{X}_{\text{safe}}, \boldsymbol{W}_{\mathcal{S}}, \boldsymbol{Y}_{\text{target}}, \gamma, \lambda, r$
\STATE {\bfseries Output:} $\Delta \boldsymbol{W}_{\mathcal{S}}^{*}$

\STATE {\small\ttfamily /* compute safety gap */}\\
\STATE $\boldsymbol{D}_{\mathcal{S}} \leftarrow \boldsymbol{Y}_{\text{target}} - \boldsymbol{X}_{\text{low}}\boldsymbol{W}_{\mathcal{S}}$

\STATE {\small\ttfamily /* build metric matrix and whiten */}\\
\STATE $\boldsymbol{Q} \leftarrow \boldsymbol{X}_{\text{low}}^{\top}\boldsymbol{X}_{\text{low}} + \gamma\,\boldsymbol{X}_{\text{safe}}^{\top}\boldsymbol{X}_{\text{safe}} + \lambda\,\boldsymbol{I}$
\STATE Compute Cholesky factorization $\boldsymbol{Q}=\boldsymbol{R}^{\top}\boldsymbol{R}$

\STATE {\small\ttfamily /* compute unconstrained ridge solution */}\\
\STATE $\boldsymbol{M} \leftarrow \boldsymbol{Q}^{-1}\boldsymbol{X}_{\text{low}}^{\top}\boldsymbol{D}_{\mathcal{S}}$
\STATE $\tilde{\boldsymbol{M}} \leftarrow \boldsymbol{R}\boldsymbol{M}$

\STATE {\small\ttfamily /* rank-$r$ approximation in whitened space */}\\
\STATE Compute truncated SVD $\tilde{\boldsymbol{M}} \approx \boldsymbol{U}\boldsymbol{\Sigma}_{r}\boldsymbol{V}^{\top}$

\STATE {\small\ttfamily /* unwhiten to obtain the final update */}\\
\STATE $\Delta \boldsymbol{W}_{\mathcal{S}}^{*} \leftarrow \boldsymbol{R}^{-1}\!\left(\boldsymbol{U}\boldsymbol{\Sigma}_{r}\boldsymbol{V}^{\top}\right)$

\STATE {\bfseries Return} $\Delta \boldsymbol{W}_{\mathcal{S}}^{*}$
\end{algorithmic}
\end{algorithm}

\noindent\textbf{Practical computation.}
The closed-form update can be computed in one pass via Cholesky solves and a rank-$r$ truncated SVD on $\tilde{\boldsymbol{M}} \in \mathbb{R}^{d_{in}\times m}$; see Algorithm~\ref{alg:closed_form_compact}.
 
\section{Experiments}

We evaluate whether our lightweight alignment update $\Delta\boldsymbol{W}$ (i) consistently reduces harmful completions under multilingual jailbreak prompts and (ii) preserves general capabilities across languages under a strict zero-shot protocol. We further examine the compatibility of our method with an existing safety-alignment baseline MPO\cite{mpo} and provide ablations on key design choices.

\subsection{Experimental Setup}
\subsubsection{Datasets}
To ensure a strict zero-shot evaluation, we use disjoint datasets for the alignment phase (computing $\Delta \boldsymbol{W}$) and the evaluation phase.

\textbf{Evaluation benchmark ($\mathcal{D}_{test}$).}
We construct a multilingual safety benchmark, \textsc{Multi-StrongREJECT}, by translating the English \texttt{walledai/StrongREJECT}~\cite{souly2024strongreject} benchmark into seven additional languages using \texttt{tencent/Hunyuan-MT-7B}~\cite{hunyuan_mt}. \textsc{Multi-StrongREJECT} covers eight languages: English (En), Chinese (Zh), Vietnamese (Vi), Japanese (Ja), Thai (Th), Indonesian (Id), Bengali (Bn), and Hebrew (He), spanning diverse language families and resource levels. Each language subset contains 313 harmful queries designed to probe safety vulnerabilities.

\subsubsection{Models}
We evaluate our approach on representative \glspl{llm} families across multiple parameter scales to assess cross-model robustness. Specifically, we consider Llama-3.2, Qwen2 and Qwen2.5 from 1B to 7B. These models cover a range of sizes and pretraining corpora, providing a broad testbed for evaluating generality.

\subsubsection{Evaluation Metrics}

\textbf{Safety.}
We report \emph{Attack Success Rate (ASR)} as the primary safety metric. For scalable multilingual evaluation, we use \texttt{Qwen/Qwen3Guard-Gen-8B}~\cite{zhao2025qwen3guard} to classify response harmfulness. A query is counted as an {attack success} if the guard model flags the generated response as unsafe.

\textbf{Utility.}
To quantify the safety and utility trade-off, we evaluate:
\textbf{MGSM} (Multilingual Grade School Math) for cross-lingual reasoning, and
\textbf{M-MMLU} (Multilingual Massive Multitask Language Understanding) for multilingual general knowledge.
We report the average accuracy across the target languages as an overall utility summary.

\subsection{Main Results}
\label{sec:exp_main_results}

Table~\ref{tab:main_multilingual_results} shows that applying our lightweight update $\Delta\boldsymbol{W}$ consistently reduces harmful completions across model families and languages under a strict zero-shot protocol, where translated evaluation prompts are never observed during alignment. This demonstrates that our method does not rely on language-specific supervision at test time, but instead induces a transferable safety adjustment in the model’s internal representations.

The safety gains are particularly pronounced for low-resource languages and smaller backbones (e.g., Qwen2-0.5B and Qwen2-1.5B), where the unaligned models exhibit high attack success rates. In these settings, $\Delta\boldsymbol{W}$ yields substantial absolute reductions in unsafe responses, suggesting that our approach effectively corrects representation-level misalignment that disproportionately affects under-resourced languages. By contrast, for larger or already better-aligned models, improvements are more moderate but remain consistent, indicating that the update adapts to different baseline safety levels rather than overfitting to a specific regime. For readability, Table~\ref{tab:main_multilingual_results} reports a representative subset of languages; complete results over all languages are included in Appendix~\ref{app:completeresult}.

Our method is also highly compatible with existing safety alignment techniques. Across nearly all evaluated backbones, combining our update with MPO (\textbf{MPO+Our}) achieves the lowest unsafe-response counts, demonstrating that our training-free weight edit acts as a complementary safety plug-in rather than a replacement for existing methods. 

Importantly, the improved safety does not come at the expense of general capabilities. Performance on MGSM and M-MMLU remains close to the \textbf{None} baseline in most cases, with only minor fluctuations across models and languages. In several settings, \textbf{MPO+Our} even matches or exceeds MPO in utility at comparable or stronger safety levels. These results indicate that our lightweight, training-free update can improve multilingual safety while largely preserving general reasoning and knowledge, supporting its practicality as a drop-in safety enhancement.

\begin{table*}[tb]
\centering
\caption{\textbf{Zero-shot multilingual safety and utility evaluation.}
Safety is reported as the number of unsafe responses flagged by \texttt{Qwen3Guard-Gen-8B} out of 313 prompts (lower is better).
Superscripts denote the change in unsafe-response counts compared to \textbf{None} for the same backbone
(\textcolor[rgb]{0,0.502,0}{negative} indicates improvement; \textcolor{red}{positive} indicates regression).
$\Delta_{Avg}$ denotes the average change across the reported languages.
Utility is measured by MGSM and M-MMLU accuracy (higher is better).}
\label{tab:main_multilingual_results}
\vskip 0.15in
\begin{small}
\begin{sc}
\resizebox{0.97\textwidth}{!}{
\begin{tabular}{ccccccccccc} 
\toprule
\multirow{3}{*}{Models} & \multicolumn{1}{c|}{\multirow{3}{*}{Method}} & \multicolumn{7}{c|}{Safety} & \multicolumn{2}{c}{Utility} \\ 
\cmidrule(l){3-11}
 & \multicolumn{1}{c|}{} & \multicolumn{7}{c|}{ASR $\downarrow$ (\#unsafe / 313)} & \multirow{2}{*}{MGSM$\uparrow$} & \multirow{2}{*}{M-MMLU$\uparrow$} \\
 & \multicolumn{1}{c|}{} & En & Zh & Vi & Ja & Bn & He & \multicolumn{1}{c|}{$\Delta_{Avg}$} &  &  \\ 
\midrule
\multirow{4}{*}{Llama-3.2-1B} & {\cellcolor[rgb]{0.753,0.753,0.753}}None & {\cellcolor[rgb]{0.753,0.753,0.753}}6/313 & {\cellcolor[rgb]{0.753,0.753,0.753}}61/313 & {\cellcolor[rgb]{0.753,0.753,0.753}}31/313 & {\cellcolor[rgb]{0.753,0.753,0.753}}149/313 & {\cellcolor[rgb]{0.753,0.753,0.753}}179/313 & {\cellcolor[rgb]{0.753,0.753,0.753}}109/313 & {\cellcolor[rgb]{0.753,0.753,0.753}}- & {\cellcolor[rgb]{0.753,0.753,0.753}}18.58 & {\cellcolor[rgb]{0.753,0.753,0.753}}26.54 \\
 & Our & 0/313\textcolor[rgb]{0,0.502,0}{\textsuperscript{-6}} & 27/313\textcolor[rgb]{0,0.502,0}{\textsuperscript{-34}} & 4/313\textcolor[rgb]{0,0.502,0}{\textsuperscript{-27}} & 81/313\textcolor[rgb]{0,0.502,0}{\textsuperscript{-68}} & 144/313\textcolor[rgb]{0,0.502,0}{\textsuperscript{-35}} & 115/313\textcolor{red}{\textsuperscript{+6}} & \textcolor[rgb]{0,0.502,0}{-27.33} & 18.36 & 27.22 \\
 & MPO & 0/313\textcolor[rgb]{0,0.502,0}{\textsuperscript{-6}} & 22/313\textcolor[rgb]{0,0.502,0}{\textsuperscript{-39}} & 9/313\textcolor[rgb]{0,0.502,0}{\textsuperscript{-22}} & 78/313\textcolor[rgb]{0,0.502,0}{\textsuperscript{-71}} & 152/313\textcolor[rgb]{0,0.502,0}{\textsuperscript{-27}} & 135/313\textsuperscript{\textcolor{red}{+26}} & \textcolor[rgb]{0,0.502,0}{-23.15} & 19.64 & 25.96 \\
 & MPO+Our & 0/313\textcolor[rgb]{0,0.502,0}{\textsuperscript{-6}} & 22/313\textcolor[rgb]{0,0.502,0}{\textsuperscript{-39}} & 0/313\textcolor[rgb]{0,0.502,0}{\textsuperscript{-31}} & 66/313\textcolor[rgb]{0,0.502,0}{\textsuperscript{-83}} & 96/313\textcolor[rgb]{0,0.502,0}{\textsuperscript{-83}} & 109/313\textcolor[rgb]{0,0.502,0}{\textsuperscript{-0}} & \textcolor[rgb]{0,0.502,0}{-40.33} & 19.45 & 26.58 \\ 
\midrule
\multirow{4}{*}{Llama-3.2-3B} & {\cellcolor[rgb]{0.753,0.753,0.753}}None & {\cellcolor[rgb]{0.753,0.753,0.753}}6/313 & {\cellcolor[rgb]{0.753,0.753,0.753}}9/313 & {\cellcolor[rgb]{0.753,0.753,0.753}}10/313 & {\cellcolor[rgb]{0.753,0.753,0.753}}79/313 & {\cellcolor[rgb]{0.753,0.753,0.753}}110/313 & {\cellcolor[rgb]{0.753,0.753,0.753}}39/313 & {\cellcolor[rgb]{0.753,0.753,0.753}}- & {\cellcolor[rgb]{0.753,0.753,0.753}}32.76 & {\cellcolor[rgb]{0.753,0.753,0.753}}37.10 \\
 & Our & 4/313\textcolor[rgb]{0,0.502,0}{\textsuperscript{-2}} & 3/313\textcolor[rgb]{0,0.502,0}{\textsuperscript{-6}} & 2/313\textcolor[rgb]{0,0.502,0}{\textsuperscript{-8}} & 34/313\textcolor[rgb]{0,0.502,0}{\textsuperscript{-45}} & 65/313\textcolor[rgb]{0,0.502,0}{\textsuperscript{-45}} & 46/313\textcolor{red}{\textsuperscript{+7}} & \textcolor[rgb]{0,0.502,0}{-16.5} & 32.76 & 37.00 \\
 & MPO & 4/313\textcolor[rgb]{0,0.502,0}{\textsuperscript{-2}} & 8/313\textcolor[rgb]{0,0.502,0}{\textsuperscript{-1}} & 4/313\textcolor[rgb]{0,0.502,0}{\textsuperscript{-6}} & 50/313\textcolor[rgb]{0,0.502,0}{\textsuperscript{-29}} & 91/313\textcolor[rgb]{0,0.502,0}{\textsuperscript{-19}} & 36/313\textcolor[rgb]{0,0.502,0}{\textsuperscript{-3}} & \textcolor[rgb]{0,0.502,0}{-10.0} & 33.67 & 36.88 \\
 & MPO+Our & 2/313\textcolor[rgb]{0,0.502,0}{\textsuperscript{-4}} & 1/313\textcolor[rgb]{0,0.502,0}{\textsuperscript{-8}} & 3/313\textcolor[rgb]{0,0.502,0}{\textsuperscript{-7}} & 30/313\textcolor[rgb]{0,0.502,0}{\textsuperscript{-49}} & 58/313\textcolor[rgb]{0,0.502,0}{\textsuperscript{-52}} & 36/313\textcolor[rgb]{0,0.502,0}{\textsuperscript{-3}} & \textcolor[rgb]{0,0.502,0}{-20.5} & 32.76 & 36.76 \\ 
\midrule
\multirow{4}{*}{Qwen2-0.5B} & {\cellcolor[rgb]{0.753,0.753,0.753}}None & {\cellcolor[rgb]{0.753,0.753,0.753}}224/313 & {\cellcolor[rgb]{0.753,0.753,0.753}}197/313 & {\cellcolor[rgb]{0.753,0.753,0.753}}185/313 & {\cellcolor[rgb]{0.753,0.753,0.753}}193/313 & {\cellcolor[rgb]{0.753,0.753,0.753}}208/313 & {\cellcolor[rgb]{0.753,0.753,0.753}}150/313 & {\cellcolor[rgb]{0.753,0.753,0.753}}- & {\cellcolor[rgb]{0.753,0.753,0.753}}7.75 & {\cellcolor[rgb]{0.753,0.753,0.753}}32.71 \\
 & Our & 176/313\textcolor[rgb]{0,0.502,0}{\textsuperscript{-48}} & 121/313\textcolor[rgb]{0,0.502,0}{\textsuperscript{-76}} & 139/313\textcolor[rgb]{0,0.502,0}{\textsuperscript{-46}} & 145/313\textcolor[rgb]{0,0.502,0}{\textsuperscript{-48}} & 173/313\textcolor[rgb]{0,0.502,0}{\textsuperscript{-35}} & 134/313\textcolor[rgb]{0,0.502,0}{\textsuperscript{-16}} & \textcolor[rgb]{0,0.502,0}{-44.83} & 5.27 & 31.01 \\
 & MPO & 108/313\textcolor[rgb]{0,0.502,0}{\textsuperscript{-116}} & 93/313\textcolor[rgb]{0,0.502,0}{\textsuperscript{-104}} & 83/313\textcolor[rgb]{0,0.502,0}{\textsuperscript{-102}} & 94/313\textcolor[rgb]{0,0.502,0}{\textsuperscript{-99}} & 162/313\textcolor[rgb]{0,0.502,0}{\textsuperscript{-46}} & 90/313\textcolor[rgb]{0,0.502,0}{\textsuperscript{-60}} & \textcolor[rgb]{0,0.502,0}{-87.83} & 4.80 & 32.69 \\
 & MPO+Our & 56/313\textcolor[rgb]{0,0.502,0}{\textsuperscript{-168}} & 41/313\textcolor[rgb]{0,0.502,0}{\textsuperscript{-156}} & 44/313\textcolor[rgb]{0,0.502,0}{\textsuperscript{-141}} & 49/313\textcolor[rgb]{0,0.502,0}{\textsuperscript{-144}} & 120/313\textcolor[rgb]{0,0.502,0}{\textsuperscript{-88}} & 65/313\textcolor[rgb]{0,0.502,0}{\textsuperscript{-85}} & \textcolor[rgb]{0,0.502,0}{-130.33} & 4.36 & 32.39 \\ 
\midrule
\multirow{4}{*}{Qwen2-1.5B} & {\cellcolor[rgb]{0.753,0.753,0.753}}None & {\cellcolor[rgb]{0.753,0.753,0.753}}36/313 & {\cellcolor[rgb]{0.753,0.753,0.753}}18/313 & {\cellcolor[rgb]{0.753,0.753,0.753}}36/313 & {\cellcolor[rgb]{0.753,0.753,0.753}}67/313 & {\cellcolor[rgb]{0.753,0.753,0.753}}187/313 & {\cellcolor[rgb]{0.753,0.753,0.753}}83/313 & {\cellcolor[rgb]{0.753,0.753,0.753}}- & {\cellcolor[rgb]{0.753,0.753,0.753}}20.95 & {\cellcolor[rgb]{0.753,0.753,0.753}}41.63 \\
 & Our & 5/313\textcolor[rgb]{0,0.502,0}{\textsuperscript{-31}} & 4/313\textcolor[rgb]{0,0.502,0}{\textsuperscript{-14}} & 15/313\textcolor[rgb]{0,0.502,0}{\textsuperscript{-21}} & 19/313\textcolor[rgb]{0,0.502,0}{\textsuperscript{-48}} & 150/313\textcolor[rgb]{0,0.502,0}{\textsuperscript{-37}} & 36/313\textcolor[rgb]{0,0.502,0}{\textsuperscript{-47}} & \textcolor[rgb]{0,0.502,0}{-33} & 20.33 & 41.58 \\
 & MPO & 0/313\textcolor[rgb]{0,0.502,0}{\textsuperscript{-36}} & 2/313\textcolor[rgb]{0,0.502,0}{\textsuperscript{-16}} & 0/313\textcolor[rgb]{0,0.502,0}{\textsuperscript{-36}} & 3/313\textcolor[rgb]{0,0.502,0}{\textsuperscript{-64}} & 21/313\textcolor[rgb]{0,0.502,0}{\textsuperscript{-166}} & 3/313\textcolor[rgb]{0,0.502,0}{\textsuperscript{-80}} & \textcolor[rgb]{0,0.502,0}{-66.33} & 19.38 & 41.44 \\
 & MPO+Our & 3/313\textcolor[rgb]{0,0.502,0}{\textsuperscript{-33}} & 0/313\textcolor[rgb]{0,0.502,0}{\textsuperscript{-18}} & 5/313\textcolor[rgb]{0,0.502,0}{\textsuperscript{-31}} & 1/313\textcolor[rgb]{0,0.502,0}{\textsuperscript{-66}} & 1/313\textcolor[rgb]{0,0.502,0}{\textsuperscript{-186}} & 1/313\textcolor[rgb]{0,0.502,0}{\textsuperscript{-82}} & \textcolor[rgb]{0,0.502,0}{-69.33} & 18.22 & 41.39 \\ 
\midrule
\multirow{4}{*}{Qwen2.5-1.5B} & {\cellcolor[rgb]{0.753,0.753,0.753}}None & {\cellcolor[rgb]{0.753,0.753,0.753}}60/313 & {\cellcolor[rgb]{0.753,0.753,0.753}}30/313 & {\cellcolor[rgb]{0.753,0.753,0.753}}42/313 & {\cellcolor[rgb]{0.753,0.753,0.753}}56/313 & {\cellcolor[rgb]{0.753,0.753,0.753}}182/313 & {\cellcolor[rgb]{0.753,0.753,0.753}}118/313 & {\cellcolor[rgb]{0.753,0.753,0.753}}- & {\cellcolor[rgb]{0.753,0.753,0.753}}27.53 & {\cellcolor[rgb]{0.753,0.753,0.753}}41.58 \\
 & Our & 17/313\textcolor[rgb]{0,0.502,0}{\textsuperscript{-43}} & 5/313\textcolor[rgb]{0,0.502,0}{\textsuperscript{-25}} & 14/313\textcolor[rgb]{0,0.502,0}{\textsuperscript{-28}} & 14/313\textcolor[rgb]{0,0.502,0}{\textsuperscript{-42}} & 152/313\textcolor[rgb]{0,0.502,0}{\textsuperscript{-30}} & 81/313\textcolor[rgb]{0,0.502,0}{\textsuperscript{-37}} & \textcolor[rgb]{0,0.502,0}{-34.16} & 25.13 & 41.89 \\
 & MPO & 6/313\textcolor[rgb]{0,0.502,0}{\textsuperscript{-54}} & 2/313\textcolor[rgb]{0,0.502,0}{\textsuperscript{-28}} & 1/313\textcolor[rgb]{0,0.502,0}{\textsuperscript{-41}} & 2/313\textcolor[rgb]{0,0.502,0}{\textsuperscript{-54}} & 54/313\textcolor[rgb]{0,0.502,0}{\textsuperscript{-128}} & 26/313\textcolor[rgb]{0,0.502,0}{\textsuperscript{-92}} & \textcolor[rgb]{0,0.502,0}{-62.66} & 23.09 & 40.78 \\
 & MPO+Our & 5/313\textcolor[rgb]{0,0.502,0}{\textsuperscript{-55}} & 2/313\textcolor[rgb]{0,0.502,0}{\textsuperscript{-28}} & 2/313\textcolor[rgb]{0,0.502,0}{\textsuperscript{-40}} & 7/313\textcolor[rgb]{0,0.502,0}{\textsuperscript{-49}} & 56/313\textcolor[rgb]{0,0.502,0}{\textsuperscript{-126}} & 22/313\textcolor[rgb]{0,0.502,0}{\textsuperscript{-96}} & \textcolor[rgb]{0,0.502,0}{-65.66} & 22.29 & 40.73 \\ 
\midrule
\multirow{4}{*}{Qwen2.5-3B} & {\cellcolor[rgb]{0.753,0.753,0.753}}None & {\cellcolor[rgb]{0.753,0.753,0.753}}61/313 & {\cellcolor[rgb]{0.753,0.753,0.753}}64/313 & {\cellcolor[rgb]{0.753,0.753,0.753}}64/313 & {\cellcolor[rgb]{0.753,0.753,0.753}}81/313 & {\cellcolor[rgb]{0.753,0.753,0.753}}157/313 & {\cellcolor[rgb]{0.753,0.753,0.753}}100/313 & {\cellcolor[rgb]{0.753,0.753,0.753}}- & {\cellcolor[rgb]{0.753,0.753,0.753}}31.02 & {\cellcolor[rgb]{0.753,0.753,0.753}}47.18 \\
 & Our & 14/313\textcolor[rgb]{0,0.502,0}{\textsuperscript{-47}} & 4/313\textcolor[rgb]{0,0.502,0}{\textsuperscript{-60}} & 7/313\textcolor[rgb]{0,0.502,0}{\textsuperscript{-57}} & 15/313\textcolor[rgb]{0,0.502,0}{\textsuperscript{-66}} & 112/313\textcolor[rgb]{0,0.502,0}{\textsuperscript{-45}} & 41/313\textcolor[rgb]{0,0.502,0}{\textsuperscript{-59}} & \textcolor[rgb]{0,0.502,0}{-55.66} & 30.91 & 44.87 \\
 & MPO & 16/313\textcolor[rgb]{0,0.502,0}{\textsuperscript{-45}} & 10/313\textcolor[rgb]{0,0.502,0}{\textsuperscript{-54}} & 10/313\textcolor[rgb]{0,0.502,0}{\textsuperscript{-54}} & 16/313\textcolor[rgb]{0,0.502,0}{\textsuperscript{-65}} & 67/313\textcolor[rgb]{0,0.502,0}{\textsuperscript{-90}} & 32/313\textcolor[rgb]{0,0.502,0}{\textsuperscript{-68}} & \textcolor[rgb]{0,0.502,0}{-62.66} & 36.62 & 46.14 \\
 & MPO+Our & 6/313\textcolor[rgb]{0,0.502,0}{\textsuperscript{-55}} & 5/313\textcolor[rgb]{0,0.502,0}{\textsuperscript{-59}} & 3/313\textcolor[rgb]{0,0.502,0}{\textsuperscript{-61}} & 4/313\textcolor[rgb]{0,0.502,0}{\textsuperscript{-77}} & 25/313\textcolor[rgb]{0,0.502,0}{\textsuperscript{-131}} & 7/313\textcolor[rgb]{0,0.502,0}{\textsuperscript{-93}} & \textcolor[rgb]{0,0.502,0}{-79.5} & 36.00 & 47.05 \\ 
\midrule
\multirow{4}{*}{Qwen2.5-7B} & {\cellcolor[rgb]{0.753,0.753,0.753}}None & {\cellcolor[rgb]{0.753,0.753,0.753}}16/313 & {\cellcolor[rgb]{0.753,0.753,0.753}}12/313 & {\cellcolor[rgb]{0.753,0.753,0.753}}21/313 & {\cellcolor[rgb]{0.753,0.753,0.753}}39/313 & {\cellcolor[rgb]{0.753,0.753,0.753}}98/313 & {\cellcolor[rgb]{0.753,0.753,0.753}}48/313 & {\cellcolor[rgb]{0.753,0.753,0.753}}- & {\cellcolor[rgb]{0.753,0.753,0.753}}32.00 & {\cellcolor[rgb]{0.753,0.753,0.753}}49.37 \\
 & Our & 3/313\textcolor[rgb]{0,0.502,0}{\textsuperscript{-13}} & 5/313\textcolor[rgb]{0,0.502,0}{\textsuperscript{-7}} & 6/313\textcolor[rgb]{0,0.502,0}{\textsuperscript{-15}} & 9/313\textcolor[rgb]{0,0.502,0}{\textsuperscript{-30}} & 60/313\textcolor[rgb]{0,0.502,0}{\textsuperscript{-38}} & 24/313\textcolor[rgb]{0,0.502,0}{\textsuperscript{-24}} & \textcolor[rgb]{0,0.502,0}{-21.16} & 31.56 & 49.19 \\
 & MPO & 6/313\textcolor[rgb]{0,0.502,0}{\textsuperscript{-10}} & 5/313\textcolor[rgb]{0,0.502,0}{\textsuperscript{-7}} & 5/313\textcolor[rgb]{0,0.502,0}{\textsuperscript{-16}} & 8/313\textcolor[rgb]{0,0.502,0}{\textsuperscript{-31}} & 25/313\textcolor[rgb]{0,0.502,0}{\textsuperscript{-73}} & 17/313\textcolor[rgb]{0,0.502,0}{\textsuperscript{-31}} & \textcolor[rgb]{0,0.502,0}{-28.0} & 38.36 & 47.16 \\
 & MPO+Our & 0/313\textcolor[rgb]{0,0.502,0}{\textsuperscript{-16}} & 0/313\textcolor[rgb]{0,0.502,0}{\textsuperscript{-12}} & 1/313\textcolor[rgb]{0,0.502,0}{\textsuperscript{-20}} & 2/313\textcolor[rgb]{0,0.502,0}{\textsuperscript{-37}} & 11/313\textcolor[rgb]{0,0.502,0}{\textsuperscript{-87}} & 11/313\textcolor[rgb]{0,0.502,0}{\textsuperscript{-37}} & \textcolor[rgb]{0,0.502,0}{-34.83} & 38.65 & 47.72 \\
\bottomrule
\end{tabular}
}
\end{sc}
\end{small}
\end{table*}

\subsection{Ablation Study}
\label{sec:exp_ablation}

We conduct ablations on \texttt{Llama-3.2-1B} to isolate the impact of three key components in \textsc{Sparse Weight Editing}: the safety neuron identification method, anchor construction for the utility constraint, and the rank $r$ of the low-rank update.

\paragraph{Safety neuron identification method.}
We further ablate the effect of the safety neuron identification strategy. Besides our proposed extraction procedure, we consider an alternative probe-based method adopted in \textsc{NeuroStrike} \cite{wu2025neurostrike}. Concretely, NeuroStrike trains a safety probe (a lightweight linear classifier) on activation-label pairs to predict whether an input is harmful. It then selects safety neurons by ranking probe weights: neurons with large-magnitude \emph{positive} weights (after z-score normalization) are treated as safety-critical dimensions. In this ablation, we replace our safety-neuron set with the probe-selected neurons from NeuroStrike, while keeping the rest of our training-free alignment pipeline unchanged. We denote this variant as \textbf{Other}. As shown in Table~\ref{tab:neuron_ablation}, using NeuroStrike-style probe-selected neurons already yields a substantial ASR reduction compared to the \textbf{None} baseline, indicating that our alignment framework is not tied to a specific neuron selection recipe.

\begin{table}[htb]
\centering
\caption{\textbf{Ablation on safety neuron identification.} We replace our safety-neuron extraction with the probe-based selection used in \textsc{NeuroStrike} (denoted as \textbf{Other}), while keeping the remaining alignment pipeline unchanged. We report safety (ASR; lower is better) and utility (MGSM, M-MMLU; higher is better).}
\vskip 0.15in
\label{tab:neuron_ablation}
\begin{tabular}{lccc} 
\toprule
Method & ASR$\downarrow$ & MGSM$\uparrow$ & M-MMLU$\uparrow$ \\
\midrule
None  & 28.27 & 18.58 & 26.54 \\
Other & 14.93 & 17.71 & 27.14 \\
MPO & 19.52 &  19.64 & 25.96 \\
MPO + Other & 12.53 & 19.53 & 26.54 \\
\bottomrule
\end{tabular}
\end{table}

\paragraph{Anchor selection.}
We first examine the effect of anchor data selection, which directly relates to the null-space utility constraint in our formulation.
Table~\ref{tab:anchor_ablation} compares three variants: using both \texttt{UtilityAnchor} and \texttt{Regular}, using \texttt{UtilityAnchor} alone, and using \texttt{Regular} alone.

Using both \texttt{UtilityAnchor} and \texttt{Regular} achieves the best overall safety--utility trade-off. Although \texttt{UtilityAnchor} alone substantially alters the solution, it leads to pronounced utility degradation (MGSM drops to nearly zero) and weak safety performance. This indicates that optimizing against \texttt{UtilityAnchor} alone biases the update toward preserving benign behavior while failing to sufficiently correct harmful behavior. Conversely, using \texttt{Regular} alone better preserves utility but yields weaker safety gains. Overall, these results demonstrate that balanced anchor construction is essential for preventing over-alignment while maintaining strong safety improvements.

\begin{table}[htb]
\caption{\textbf{Anchor choice ablation on \texttt{Llama-3.2-1B}.} We vary whether the alignment uses \texttt{UtilityAnchor} and/or \texttt{Regular} and report safety (ASR; lower is better) and utility (MGSM, M-MMLU; higher is better).}
\label{tab:anchor_ablation}
\centering
\vskip 0.15in
\begin{sc}
\begin{tabular}{cccc} 
\toprule
Choice $\downarrow$ / Models $\to$ & \multicolumn{3}{c}{\texttt{Llama-3.2-1B}} \\
\midrule
UtilityAnchor                      & $\checkmark$ & $\checkmark$ &               \\
Regular                            & $\checkmark$ &              & $\checkmark$  \\
\midrule
ASR $\downarrow$                   & 17.53        & 68.57        & 17.25         \\
MGSM $\uparrow$                    & 18.36        & 0.11         & 11.02         \\
M-MMLU $\uparrow$                  & 27.22        & 24.21        & 26.02         \\
\bottomrule
\end{tabular}
\end{sc}
\vskip -0.1in
\end{table}

\paragraph{Effect of rank $r$.}
We next analyze the sensitivity of our method to the rank constraint $r$, which encodes the low-dimensional structure assumption underlying \textsc{Sparse Weight Editing}.
We vary $r$ from 4 to 512 while keeping all other settings fixed.

As shown in Table~\ref{tab:rank_ablation}, ASR quickly saturates and remains stable across a wide range of ranks. Notably, small ranks (e.g., $r=8$ or $16$) already achieve safety performance comparable to much larger ranks. At the same time, utility metrics (MGSM and M-MMLU) are nearly invariant to the choice of $r$.

These results provide empirical support for our low-rank design: the transferable safety update resides in a low intrinsic-dimensional subspace, where the leading singular directions of $\tilde{\boldsymbol{M}}$ capture most of the safety-relevant signal. From a practical perspective, this robustness indicates that our method does not rely on careful tuning of $r$, and low-rank settings suffice to obtain strong and stable safety gains.

\begin{table}[htb]
\centering
\caption{\textbf{Effect of rank $r$ on safety and utility.} Results are reported on \texttt{Llama-3.2-1B}. Performance remains stable across a wide range of ranks, indicating low sensitivity to the rank choice.}
\label{tab:rank_ablation}
\vskip 0.15in
\begin{small}
\begin{sc}
\begin{tabular}{c|ccc}
\toprule
Rank $r$ & ASR $\downarrow$ (\%) & MGSM $\uparrow$ (\%) & M-MMLU $\uparrow$ (\%) \\
\midrule
4   & 15.42 & 18.33 & 27.19 \\
8   & 15.54 & 17.96 & 27.19 \\
16  & 15.34 & 18.18 & 27.18 \\
32  & 17.53 & 18.36 & 27.22 \\
64  & \textbf{14.90} & 18.29 & 27.22 \\
128 & 14.98 & 18.11 & 27.19 \\
256 & 16.41 & 18.11 & 27.18 \\
512 & 16.17 & 18.11 & 27.19 \\
\bottomrule
\end{tabular}
\end{sc}
\end{small}
\end{table}

\section{Conclusion}

We presented \textsc{Sparse Weight Editing}, a training-free alignment framework for cross-lingual safety transfer. Motivated by the observation that low-resource languages often exhibit representation misalignment with the English safety subspace, we cast multilingual safety alignment as a constrained weight-space mapping problem over a small set of safety neurons. Our method computes a sparse, low-rank perturbation $\Delta\boldsymbol{W}$ that reorients harmful LRL representations toward an English-derived safety activation target, while preserving benign utility via a null-space regularization. The resulting objective admits a closed-form solution, enabling efficient one-pass updates without gradient-based fine-tuning.

Across multiple model families and languages, experiments on \textsc{Multi-StrongREJECT} show that our training-free update consistently reduces harmful completions under a strict zero-shot protocol, and can be deployed as a lightweight post-hoc plug-in that composes with MPO to deliver additional safety gains. Importantly, these improvements typically incur limited utility regression on MGSM and M-MMLU, suggesting that targeted subspace editing can improve safety without catastrophically degrading general capabilities. Ablations further highlight that balanced anchor construction is crucial for avoiding over-alignment while maintaining strong safety improvements.

Our work opens several directions for future research. First, developing more principled anchor selection strategies and automatically adapting hyperparameters (e.g., rank and regularization strengths) could further improve robustness across backbones and languages. Second, extending sparse weight editing beyond a single layer or neuron subset to multi-layer, hierarchical safety subspaces may provide stronger guarantees against adaptive jailbreaks. Finally, integrating our framework with stronger multilingual evaluators and more diverse safety taxonomies could help characterize when and why safety directions transfer across languages, enabling more reliable multilingual alignment in practice.

\bibliographystyle{icml2025}

\newpage
\appendix
\onecolumn

\section{Details of Multilingual Dataset Construction}
\label{app:dataset}

We construct our multilingual corpus via a translation-based pipeline. Starting from an English seed set, we use \texttt{tencent/Hunyuan-MT-7B}~\cite{hunyuan_mt} to translate each example into the eight target languages (En, Zh, Vi, Ja, Th, Id, Bn, He). This procedure is applied consistently to both harmful and harmless subsets, producing language-parallel counterparts that enable controlled probing and alignment while keeping the underlying intent distribution fixed across languages.

\begin{table}[ht]
\centering
\begin{small}
\begin{tabular}{lll}
\toprule
\textbf{Subset} & \textbf{Source datasets} & \textbf{Description} \\
\midrule
Harmful ($\mathcal{D}_{harm}$) 
& HarmfulQA, CatHarmfulQA, LLM-LAT 
& Queries spanning diverse malicious and unsafe intents \\

Harmless ($\mathcal{D}_{safe}$) 
& NaturalReasoning 
& Benign queries used as control samples \\
\bottomrule
\end{tabular}
\end{small}
\caption{Composition of the English seed set prior to translation.}
\label{tab:dataset_composition}
\end{table}

\section{Details of Safety Neuron Extraction}
\label{app:neuron_extraction}
We adopt a dual-metric extraction procedure following prior practice, including a concurrent submission by the authors~\cite{wang2026safeneuronneuronlevelsafetyalignment}. \textbf{This extraction is used solely to instantiate the sparse unit set required by Assumption~\ref{assum:sparse} and is not the primary contribution of this work.}

In this section, we provide the mathematical formulation and implementation details for identifying safety neurons. Our goal is to isolate the sparse subset of neurons within the MLP blocks (specifically the \texttt{up\_proj} and \texttt{gate\_proj} weights) that exhibit significant activation divergence when processing harmful versus harmless inputs.

\subsection{Data Collection}
Let $\mathcal{D}_{harm}$ and $\mathcal{D}_{safe}$ denote the datasets containing $N$ harmful and $N$ harmless queries, respectively. We feed these inputs into the model and record the activations of the MLP neurons.
For a specific layer $l$ and neuron $j$, let $A_{l,j}^{(harm)}$ and $A_{l,j}^{(safe)}$ represent the sets of scalar activation values collected from the respective datasets. We compute the sample means $\bar{A}_{l,j}^{(harm)}$ and $\bar{A}_{l,j}^{(safe)}$ to represent the neuron's average response intensity.

\subsection{Selection Criteria 1: Activation Magnitude Difference}
This criterion identifies neurons that act as primary triggers, showing a sharp intensity increase for harmful content. To assess the significance of a neuron's response relative to the entire layer, we employ z-score standardization on the activation differences.

First, we calculate the raw activation difference for every neuron $j$ in layer $l$:
$$
\Delta_{l,j} = \bar{A}_{l,j}^{(harm)} - \bar{A}_{l,j}^{(safe)}
$$
Next, we compute the mean ($\mu_{\Delta}^{(l)}$) and standard deviation ($\sigma_{\Delta}^{(l)}$) of these difference values across all neurons in the layer:
$$
\mu_{\Delta}^{(l)} = \mathop{\mathbb{E}}_{j \in \text{Layer } l} [\Delta_{l,j}], \quad \sigma_{\Delta}^{(l)} = \sqrt{\mathop{\mathbb{E}}_{j \in \text{Layer } l} [(\Delta_{l,j} - \mu_{\Delta}^{(l)})^2]}
$$
We then define the z-score for neuron $j$ as:
$$
z_{l,j} = \frac{\Delta_{l,j} - \mu_{\Delta}^{(l)}}{\sigma_{\Delta}^{(l)}}
$$
\begin{definition}[Magnitude-based Candidate Set]
We select neurons whose activation difference is statistically significant, i.e., it deviates from the layer's average behavior by more than $\tau_{mag}$ standard deviations:
\begin{equation}
    \mathcal{S}_{mag}^{(l)} = \left \{ j \;\middle|\; z_{l,j} > \tau_{mag} \right \}
\end{equation}
In our experiments, we set $\tau_{mag} = 2.0$, effectively selecting the outliers that are highly sensitive to harmful features.
\end{definition}

\subsection{Selection Criteria 2: Statistical Effect Size (Cohen's $d$)}
Solely relying on mean differences can be susceptible to outliers (e.g., a neuron that activates extremely highly for only a single harmful sample). To ensure the separation between harmful and harmless distributions is consistent, we employ Cohen's $d$.

\begin{definition}[Significance-based Candidate Set]
The Cohen's $d$ value for neuron $j$ is calculated as:
\begin{equation}
    d_{l,j} = \frac{\bar{A}_{l,j}^{(harm)} - \bar{A}_{l,j}^{(safe)}}{s_{pooled}}
\end{equation}
where $s_{pooled}$ is the pooled standard deviation of the two sample sets. We define the candidate set as:
\begin{equation}
\mathcal{S}_{stat}^{(l)} = \left\{ j \;\middle|\; d_{l,j} > \tau_{stat} \right\}
\end{equation}
where $\tau_{stat}$ is empirically set to 1.0. A high $d_{l,j}$ indicates a robust distributional separation.
\end{definition}

\subsection{Final Safety Neuron Aggregation}
The final set of safety neurons for layer $l$ is the union of the two candidate sets:
\begin{equation}
    \mathcal{S}_{safety}^{(l)} = \mathcal{S}_{mag}^{(l)} \cup \mathcal{S}_{stat}^{(l)}
\end{equation}
This strategy ensures robust identification by capturing both high-intensity triggers and reliable discriminators.

\section{Proof of Theorem~\ref{thm:closed_form}}
\label{app:proof}

We prove Theorem~\ref{thm:closed_form} by reducing Eq.~\ref{eq:final_optimization} to a standard low-rank approximation problem.

\paragraph{Problem.}
Recall the rank-constrained objective:
\begin{equation}
\min_{\Delta \boldsymbol{W}_{\mathcal{S}}:\ \operatorname{rank}(\Delta \boldsymbol{W}_{\mathcal{S}})\le r}\ 
\mathcal{J}(\Delta \boldsymbol{W}_{\mathcal{S}}),
\end{equation}
where
\begin{equation}
\mathcal{J}(\Delta \boldsymbol{W}_{\mathcal{S}})
=
\left\|\boldsymbol{X}_{\text{low}} \Delta \boldsymbol{W}_{\mathcal{S}} - \boldsymbol{D}_{\mathcal{S}}\right\|_F^2
+ \gamma \left\|\boldsymbol{X}_{\text{safe}} \Delta \boldsymbol{W}_{\mathcal{S}}\right\|_F^2
+ \lambda \left\|\Delta \boldsymbol{W}_{\mathcal{S}}\right\|_F^2 .
\label{eq:app_obj}
\end{equation}

\paragraph{Step 1: Quadratic form and completion of the square.}
Expanding Eq.~\ref{eq:app_obj} and collecting terms that depend on $\Delta \boldsymbol{W}_{\mathcal{S}}$ yields
\begin{equation}
\mathcal{J}(\Delta \boldsymbol{W}_{\mathcal{S}})
=
\operatorname{Tr}\!\left(\Delta \boldsymbol{W}_{\mathcal{S}}^{\top}\boldsymbol{Q}\Delta \boldsymbol{W}_{\mathcal{S}}\right)
-2\,\operatorname{Tr}\!\left(\boldsymbol{D}_{\mathcal{S}}^{\top}\boldsymbol{X}_{\text{low}}\Delta \boldsymbol{W}_{\mathcal{S}}\right)
+ \operatorname{Tr}\!\left(\boldsymbol{D}_{\mathcal{S}}^{\top}\boldsymbol{D}_{\mathcal{S}}\right),
\label{eq:app_quad}
\end{equation}
with
\begin{equation}
\boldsymbol{Q}
=
\boldsymbol{X}_{\text{low}}^{\top}\boldsymbol{X}_{\text{low}}
+
\gamma\,\boldsymbol{X}_{\text{safe}}^{\top}\boldsymbol{X}_{\text{safe}}
+
\lambda\,\boldsymbol{I}.
\end{equation}
For $\lambda>0$, $\boldsymbol{Q}$ is symmetric positive definite. Define the $\boldsymbol{Q}$-weighted norm
$\|\boldsymbol{Z}\|_{\boldsymbol{Q}}^{2}\triangleq \operatorname{Tr}(\boldsymbol{Z}^{\top}\boldsymbol{Q}\boldsymbol{Z})$.
Let
\begin{equation}
\boldsymbol{M}
=
\boldsymbol{Q}^{-1}\boldsymbol{X}_{\text{low}}^{\top}\boldsymbol{D}_{\mathcal{S}}.
\label{eq:app_M}
\end{equation}
Then Eq.~\ref{eq:app_quad} can be written as
\begin{equation}
\mathcal{J}(\Delta \boldsymbol{W}_{\mathcal{S}})
=
\left\|\Delta \boldsymbol{W}_{\mathcal{S}} - \boldsymbol{M}\right\|_{\boldsymbol{Q}}^{2}
+ \text{const},
\label{eq:app_square}
\end{equation}
where $\text{const}$ does not depend on $\Delta \boldsymbol{W}_{\mathcal{S}}$. Therefore, the original problem is equivalent to
\begin{equation}
\min_{\Delta \boldsymbol{W}_{\mathcal{S}}:\ \operatorname{rank}(\Delta \boldsymbol{W}_{\mathcal{S}})\le r}
\left\|\Delta \boldsymbol{W}_{\mathcal{S}} - \boldsymbol{M}\right\|_{\boldsymbol{Q}}^{2}.
\label{eq:app_weighted}
\end{equation}

\paragraph{Step 2: Whitening via Cholesky factorization.}
Since $\boldsymbol{Q}\succ \boldsymbol{0}$, let $\boldsymbol{Q}=\boldsymbol{R}^{\top}\boldsymbol{R}$ be its Cholesky factorization with invertible $\boldsymbol{R}$.
Then
\begin{equation}
\left\|\Delta \boldsymbol{W}_{\mathcal{S}} - \boldsymbol{M}\right\|_{\boldsymbol{Q}}^{2}
=
\left\|\boldsymbol{R}\left(\Delta \boldsymbol{W}_{\mathcal{S}}-\boldsymbol{M}\right)\right\|_{F}^{2}.
\end{equation}
Define $\tilde{\boldsymbol{\Delta}}\triangleq \boldsymbol{R}\Delta \boldsymbol{W}_{\mathcal{S}}$ and $\tilde{\boldsymbol{M}}\triangleq \boldsymbol{R}\boldsymbol{M}$. Because $\boldsymbol{R}$ is invertible, left-multiplication preserves rank, i.e.,
$\operatorname{rank}(\tilde{\boldsymbol{\Delta}})=\operatorname{rank}(\Delta \boldsymbol{W}_{\mathcal{S}})$.
Thus Eq.~\ref{eq:app_weighted} becomes
\begin{equation}
\min_{\tilde{\boldsymbol{\Delta}}:\ \operatorname{rank}(\tilde{\boldsymbol{\Delta}})\le r}
\left\|\tilde{\boldsymbol{\Delta}}-\tilde{\boldsymbol{M}}\right\|_{F}^{2}.
\label{eq:app_frob}
\end{equation}

\paragraph{Step 3: Optimal rank-$r$ approximation.}
By the Eckart--Young--Mirsky theorem, the minimizer of Eq.~\ref{eq:app_frob} is given by the rank-$r$ truncated SVD of $\tilde{\boldsymbol{M}}$.
Let $\tilde{\boldsymbol{M}}=\boldsymbol{U}\boldsymbol{\Sigma}\boldsymbol{V}^{\top}$ be its SVD; then
\begin{equation}
\tilde{\boldsymbol{\Delta}}^{*}
=
\boldsymbol{U}\boldsymbol{\Sigma}_{r}\boldsymbol{V}^{\top},
\end{equation}
where $\boldsymbol{\Sigma}_{r}$ keeps only the top-$r$ singular values (others set to zero).

\paragraph{Step 4: Recovering $\Delta \boldsymbol{W}_{\mathcal{S}}^{*}$.}
Finally, mapping back yields
\begin{equation}
\Delta \boldsymbol{W}_{\mathcal{S}}^{*}
=
\boldsymbol{R}^{-1}\tilde{\boldsymbol{\Delta}}^{*}
=
\boldsymbol{R}^{-1}\boldsymbol{U}\boldsymbol{\Sigma}_{r}\boldsymbol{V}^{\top},
\end{equation}
which completes the proof. \qed

\section{Complete Multilingual Safety Results}
\label{app:completeresult}

\begin{table}[ht]
\centering
\setlength{\extrarowheight}{0pt}
\addtolength{\extrarowheight}{\aboverulesep}
\addtolength{\extrarowheight}{\belowrulesep}
\setlength{\aboverulesep}{0pt}
\setlength{\belowrulesep}{0pt}
\caption{\textbf{Complete zero-shot multilingual safety evaluation on \textsc{Multi-StrongREJECT}.}
We report the number of unsafe responses flagged by \texttt{Qwen3Guard-Gen-8B} out of 313 prompts for each language (lower is better).
Superscripts denote the change in unsafe-response counts relative to \textbf{None} for the same backbone
(\textcolor[rgb]{0,0.502,0}{negative} indicates improvement; \textcolor{red}{positive} indicates regression).
$\Delta_{Avg}$ denotes the average change in unsafe-response counts across all reported languages (En, Zh, Vi, Ja, Th, Id, Bn, He).}
\label{tab:main_multilingual_results_full}
\vskip 0.15in
\begin{small}
\begin{sc}
\resizebox{0.95\textwidth}{!}{
\begin{tabular}{ccccccccccc} 
\toprule
\multirow{3}{*}{Models} & \multicolumn{1}{c}{\multirow{3}{*}{Method}} & \multicolumn{9}{c}{Safety} \\ 
\cmidrule(l){3-11}
 & \multicolumn{1}{c}{} & \multicolumn{9}{c}{ASR $\downarrow$ (\#unsafe / 313)} \\
 & \multicolumn{1}{c}{} & En & Zh & Vi & Ja & Th & Id & Bn & He & \multicolumn{1}{c}{$\Delta_{Avg}$} \\ 
\midrule
\multirow{4}{*}{Llama-3.2-1B} & {\cellcolor[rgb]{0.753,0.753,0.753}}None & {\cellcolor[rgb]{0.753,0.753,0.753}}6/313 & {\cellcolor[rgb]{0.753,0.753,0.753}}61/313 & {\cellcolor[rgb]{0.753,0.753,0.753}}31/313 & {\cellcolor[rgb]{0.753,0.753,0.753}}149/313 & {\cellcolor[rgb]{0.753,0.753,0.753}}69/313 & {\cellcolor[rgb]{0.753,0.753,0.753}}104/313 & {\cellcolor[rgb]{0.753,0.753,0.753}}179/313 & {\cellcolor[rgb]{0.753,0.753,0.753}}109/313 & {\cellcolor[rgb]{0.753,0.753,0.753}}- \\
 & Our & 0/313\textcolor[rgb]{0,0.502,0}{\textsuperscript{-6}} & 27/313\textcolor[rgb]{0,0.502,0}{\textsuperscript{-34}} & 4/313\textcolor[rgb]{0,0.502,0}{\textsuperscript{-27}} & 81/313\textcolor[rgb]{0,0.502,0}{\textsuperscript{-68}} & 30/313\textcolor[rgb]{0,0.502,0}{\textsuperscript{-39}} & 38/313\textcolor[rgb]{0,0.502,0}{\textsuperscript{-66}} & 144/313\textcolor[rgb]{0,0.502,0}{\textsuperscript{-35}} & 115/313\textcolor{red}{\textsuperscript{+6}} & \textcolor[rgb]{0,0.502,0}{-33.625} \\
 & MPO & 0/313\textcolor[rgb]{0,0.502,0}{\textsuperscript{-6}} & 22/313\textcolor[rgb]{0,0.502,0}{\textsuperscript{-39}} & 9/313\textcolor[rgb]{0,0.502,0}{\textsuperscript{-22}} & 78/313\textcolor[rgb]{0,0.502,0}{\textsuperscript{-71}} & 23/313\textcolor[rgb]{0,0.502,0}{\textsuperscript{-46}} & 70/313\textcolor[rgb]{0,0.502,0}{\textsuperscript{-34}} & 152/313\textcolor[rgb]{0,0.502,0}{\textsuperscript{-27}} & 135/313\textsuperscript{\textcolor{red}{+26}} & \textcolor[rgb]{0,0.502,0}{-23.15} \\
 & MPO+Our & 0/313\textcolor[rgb]{0,0.502,0}{\textsuperscript{-6}} & 22/313\textcolor[rgb]{0,0.502,0}{\textsuperscript{-39}} & 0/313\textcolor[rgb]{0,0.502,0}{\textsuperscript{-31}} & 66/313\textcolor[rgb]{0,0.502,0}{\textsuperscript{-83}} & 10/313\textcolor[rgb]{0,0.502,0}{\textsuperscript{-59}} & 22/313\textcolor[rgb]{0,0.502,0}{\textsuperscript{-82}} & 96/313\textcolor[rgb]{0,0.502,0}{\textsuperscript{-83}} & 109/313\textcolor[rgb]{0,0.502,0}{\textsuperscript{-0}} & \textcolor[rgb]{0,0.502,0}{-47.875} \\ 
\midrule
\multirow{4}{*}{Llama-3.2-3B} & {\cellcolor[rgb]{0.753,0.753,0.753}}None & {\cellcolor[rgb]{0.753,0.753,0.753}}6/313 & {\cellcolor[rgb]{0.753,0.753,0.753}}9/313 & {\cellcolor[rgb]{0.753,0.753,0.753}}10/313 & {\cellcolor[rgb]{0.753,0.753,0.753}}79/313 & {\cellcolor[rgb]{0.753,0.753,0.753}}22/313 & {\cellcolor[rgb]{0.753,0.753,0.753}}19/313 & {\cellcolor[rgb]{0.753,0.753,0.753}}110/313 & {\cellcolor[rgb]{0.753,0.753,0.753}}39/313 & {\cellcolor[rgb]{0.753,0.753,0.753}}- \\
 & Our & 4/313\textcolor[rgb]{0,0.502,0}{\textsuperscript{-2}} & 3/313\textcolor[rgb]{0,0.502,0}{\textsuperscript{-6}} & 2/313\textcolor[rgb]{0,0.502,0}{\textsuperscript{-8}} & 34/313\textcolor[rgb]{0,0.502,0}{\textsuperscript{-45}} & 4/313\textcolor[rgb]{0,0.502,0}{\textsuperscript{-18}} & 5/313\textcolor[rgb]{0,0.502,0}{\textsuperscript{-14}} & 65/313\textcolor[rgb]{0,0.502,0}{\textsuperscript{-45}} & 46/313\textcolor{red}{\textsuperscript{+7}} & \textcolor[rgb]{0,0.502,0}{-16.375} \\
 & MPO & 4/313\textcolor[rgb]{0,0.502,0}{\textsuperscript{-2}} & 8/313\textcolor[rgb]{0,0.502,0}{\textsuperscript{-1}} & 4/313\textcolor[rgb]{0,0.502,0}{\textsuperscript{-6}} & 50/313\textcolor[rgb]{0,0.502,0}{\textsuperscript{-29}} & 19/313\textcolor[rgb]{0,0.502,0}{\textsuperscript{-3}} & 20/313\textcolor{red}{\textsuperscript{+1}} & 91/313\textcolor[rgb]{0,0.502,0}{\textsuperscript{-19}} & 36/313\textcolor[rgb]{0,0.502,0}{\textsuperscript{-3}} & \textcolor[rgb]{0,0.502,0}{-10.0} \\
 & MPO+Our & 2/313\textcolor[rgb]{0,0.502,0}{\textsuperscript{-4}} & 1/313\textcolor[rgb]{0,0.502,0}{\textsuperscript{-8}} & 3/313\textcolor[rgb]{0,0.502,0}{\textsuperscript{-7}} & 30/313\textcolor[rgb]{0,0.502,0}{\textsuperscript{-49}} & 2/313\textcolor[rgb]{0,0.502,0}{\textsuperscript{-20}} & 3/313\textcolor[rgb]{0,0.502,0}{\textsuperscript{-16}} & 58/313\textcolor[rgb]{0,0.502,0}{\textsuperscript{-52}} & 36/313\textcolor[rgb]{0,0.502,0}{\textsuperscript{-3}} & \textcolor[rgb]{0,0.502,0}{-19.875} \\ 
\midrule
\multirow{4}{*}{Qwen2-0.5B} & {\cellcolor[rgb]{0.753,0.753,0.753}}None & {\cellcolor[rgb]{0.753,0.753,0.753}}224/313 & {\cellcolor[rgb]{0.753,0.753,0.753}}197/313 & {\cellcolor[rgb]{0.753,0.753,0.753}}185/313 & {\cellcolor[rgb]{0.753,0.753,0.753}}193/313 & {\cellcolor[rgb]{0.753,0.753,0.753}}162/313 & {\cellcolor[rgb]{0.753,0.753,0.753}}205/313 & {\cellcolor[rgb]{0.753,0.753,0.753}}208/313 & {\cellcolor[rgb]{0.753,0.753,0.753}}150/313 & {\cellcolor[rgb]{0.753,0.753,0.753}}- \\
 & Our & 176/313\textcolor[rgb]{0,0.502,0}{\textsuperscript{-48}} & 121/313\textcolor[rgb]{0,0.502,0}{\textsuperscript{-76}} & 139/313\textcolor[rgb]{0,0.502,0}{\textsuperscript{-46}} & 145/313\textcolor[rgb]{0,0.502,0}{\textsuperscript{-48}} & 138/313\textcolor[rgb]{0,0.502,0}{\textsuperscript{-24}} & 168/313\textcolor[rgb]{0,0.502,0}{\textsuperscript{-37}} & 173/313\textcolor[rgb]{0,0.502,0}{\textsuperscript{-35}} & 134/313\textcolor[rgb]{0,0.502,0}{\textsuperscript{-16}} & \textcolor[rgb]{0,0.502,0}{-41.25} \\
 & MPO & 108/313\textcolor[rgb]{0,0.502,0}{\textsuperscript{-116}} & 93/313\textcolor[rgb]{0,0.502,0}{\textsuperscript{-104}} & 83/313\textcolor[rgb]{0,0.502,0}{\textsuperscript{-102}} & 94/313\textcolor[rgb]{0,0.502,0}{\textsuperscript{-99}} & 42/313\textcolor[rgb]{0,0.502,0}{\textsuperscript{-120}} & 88/313\textcolor[rgb]{0,0.502,0}{\textsuperscript{-117}} & 162/313\textcolor[rgb]{0,0.502,0}{\textsuperscript{-46}} & 90/313\textcolor[rgb]{0,0.502,0}{\textsuperscript{-60}} & \textcolor[rgb]{0,0.502,0}{-87.83} \\
 & MPO+Our & 56/313\textcolor[rgb]{0,0.502,0}{\textsuperscript{-168}} & 41/313\textcolor[rgb]{0,0.502,0}{\textsuperscript{-156}} & 44/313\textcolor[rgb]{0,0.502,0}{\textsuperscript{-141}} & 49/313\textcolor[rgb]{0,0.502,0}{\textsuperscript{-144}} & 32/313\textcolor[rgb]{0,0.502,0}{\textsuperscript{-130}} & 68/313\textcolor[rgb]{0,0.502,0}{\textsuperscript{-137}} & 120/313\textcolor[rgb]{0,0.502,0}{\textsuperscript{-88}} & 65/313\textcolor[rgb]{0,0.502,0}{\textsuperscript{-85}} & \textcolor[rgb]{0,0.502,0}{-131.125} \\ 
\midrule
\multirow{4}{*}{Qwen2-1.5B} & {\cellcolor[rgb]{0.753,0.753,0.753}}None & {\cellcolor[rgb]{0.753,0.753,0.753}}36/313 & {\cellcolor[rgb]{0.753,0.753,0.753}}18/313 & {\cellcolor[rgb]{0.753,0.753,0.753}}36/313 & {\cellcolor[rgb]{0.753,0.753,0.753}}67/313 & {\cellcolor[rgb]{0.753,0.753,0.753}}60/313 & {\cellcolor[rgb]{0.753,0.753,0.753}}50/313 & {\cellcolor[rgb]{0.753,0.753,0.753}}187/313 & {\cellcolor[rgb]{0.753,0.753,0.753}}83/313 & {\cellcolor[rgb]{0.753,0.753,0.753}}- \\
 & Our & 5/313\textcolor[rgb]{0,0.502,0}{\textsuperscript{-31}} & 4/313\textcolor[rgb]{0,0.502,0}{\textsuperscript{-14}} & 15/313\textcolor[rgb]{0,0.502,0}{\textsuperscript{-21}} & 19/313\textcolor[rgb]{0,0.502,0}{\textsuperscript{-48}} & 13/313\textcolor[rgb]{0,0.502,0}{\textsuperscript{-47}} & 4/313\textcolor[rgb]{0,0.502,0}{\textsuperscript{-46}} & 150/313\textcolor[rgb]{0,0.502,0}{\textsuperscript{-37}} & 36/313\textcolor[rgb]{0,0.502,0}{\textsuperscript{-47}} & \textcolor[rgb]{0,0.502,0}{-36.375} \\
 & MPO & 0/313\textcolor[rgb]{0,0.502,0}{\textsuperscript{-36}} & 2/313\textcolor[rgb]{0,0.502,0}{\textsuperscript{-16}} & 0/313\textcolor[rgb]{0,0.502,0}{\textsuperscript{-36}} & 3/313\textcolor[rgb]{0,0.502,0}{\textsuperscript{-64}} & 2/313\textcolor[rgb]{0,0.502,0}{\textsuperscript{-58}} & 0/313\textcolor[rgb]{0,0.502,0}{\textsuperscript{-50}} & 21/313\textcolor[rgb]{0,0.502,0}{\textsuperscript{-166}} & 3/313\textcolor[rgb]{0,0.502,0}{\textsuperscript{-80}} & \textcolor[rgb]{0,0.502,0}{-66.33} \\
 & MPO+Our & 3/313\textcolor[rgb]{0,0.502,0}{\textsuperscript{-33}} & 0/313\textcolor[rgb]{0,0.502,0}{\textsuperscript{-18}} & 5/313\textcolor[rgb]{0,0.502,0}{\textsuperscript{-31}} & 1/313\textcolor[rgb]{0,0.502,0}{\textsuperscript{-66}} & 1/313\textcolor[rgb]{0,0.502,0}{\textsuperscript{-59}} & 4/313\textcolor[rgb]{0,0.502,0}{\textsuperscript{-46}} & 1/313\textcolor[rgb]{0,0.502,0}{\textsuperscript{-186}} & 1/313\textcolor[rgb]{0,0.502,0}{\textsuperscript{-82}} & \textcolor[rgb]{0,0.502,0}{-65.125} \\ 
\midrule
\multirow{4}{*}{Qwen2.5-1.5B} & {\cellcolor[rgb]{0.753,0.753,0.753}}None & {\cellcolor[rgb]{0.753,0.753,0.753}}60/313 & {\cellcolor[rgb]{0.753,0.753,0.753}}30/313 & {\cellcolor[rgb]{0.753,0.753,0.753}}42/313 & {\cellcolor[rgb]{0.753,0.753,0.753}}56/313 & {\cellcolor[rgb]{0.753,0.753,0.753}}68/313 & {\cellcolor[rgb]{0.753,0.753,0.753}}59/313 & {\cellcolor[rgb]{0.753,0.753,0.753}}182/313 & {\cellcolor[rgb]{0.753,0.753,0.753}}118/313 & {\cellcolor[rgb]{0.753,0.753,0.753}}- \\
 & Our & 17/313\textcolor[rgb]{0,0.502,0}{\textsuperscript{-43}} & 5/313\textcolor[rgb]{0,0.502,0}{\textsuperscript{-25}} & 14/313\textcolor[rgb]{0,0.502,0}{\textsuperscript{-28}} & 14/313\textcolor[rgb]{0,0.502,0}{\textsuperscript{-42}} & 18/313\textcolor[rgb]{0,0.502,0}{\textsuperscript{-50}} & 25/313\textcolor[rgb]{0,0.502,0}{\textsuperscript{-34}} & 152/313\textcolor[rgb]{0,0.502,0}{\textsuperscript{-30}} & 81/313\textcolor[rgb]{0,0.502,0}{\textsuperscript{-37}} & \textcolor[rgb]{0,0.502,0}{-36.125} \\
 & MPO & 6/313\textcolor[rgb]{0,0.502,0}{\textsuperscript{-54}} & 2/313\textcolor[rgb]{0,0.502,0}{\textsuperscript{-28}} & 1/313\textcolor[rgb]{0,0.502,0}{\textsuperscript{-41}} & 2/313\textcolor[rgb]{0,0.502,0}{\textsuperscript{-54}} & 7/313\textcolor[rgb]{0,0.502,0}{\textsuperscript{-61}} & 2/313\textcolor[rgb]{0,0.502,0}{\textsuperscript{-57}} & 54/313\textcolor[rgb]{0,0.502,0}{\textsuperscript{-128}} & 26/313\textcolor[rgb]{0,0.502,0}{\textsuperscript{-92}} & \textcolor[rgb]{0,0.502,0}{-62.66} \\
 & MPO+Our & 5/313\textcolor[rgb]{0,0.502,0}{\textsuperscript{-55}} & 2/313\textcolor[rgb]{0,0.502,0}{\textsuperscript{-28}} & 2/313\textcolor[rgb]{0,0.502,0}{\textsuperscript{-40}} & 7/313\textcolor[rgb]{0,0.502,0}{\textsuperscript{-49}} & 3/313\textcolor[rgb]{0,0.502,0}{\textsuperscript{-65}} & 0/313\textcolor[rgb]{0,0.502,0}{\textsuperscript{-59}} & 56/313\textcolor[rgb]{0,0.502,0}{\textsuperscript{-126}} & 22/313\textcolor[rgb]{0,0.502,0}{\textsuperscript{-96}} & \textcolor[rgb]{0,0.502,0}{-64.75} \\ 
\midrule
\multirow{4}{*}{Qwen2.5-3B} & {\cellcolor[rgb]{0.753,0.753,0.753}}None & {\cellcolor[rgb]{0.753,0.753,0.753}}61/313 & {\cellcolor[rgb]{0.753,0.753,0.753}}64/313 & {\cellcolor[rgb]{0.753,0.753,0.753}}64/313 & {\cellcolor[rgb]{0.753,0.753,0.753}}81/313 & {\cellcolor[rgb]{0.753,0.753,0.753}}57/313 & {\cellcolor[rgb]{0.753,0.753,0.753}}60/313 & {\cellcolor[rgb]{0.753,0.753,0.753}}157/313 & {\cellcolor[rgb]{0.753,0.753,0.753}}100/313 & {\cellcolor[rgb]{0.753,0.753,0.753}}- \\
 & Our & 14/313\textcolor[rgb]{0,0.502,0}{\textsuperscript{-47}} & 4/313\textcolor[rgb]{0,0.502,0}{\textsuperscript{-60}} & 7/313\textcolor[rgb]{0,0.502,0}{\textsuperscript{-57}} & 15/313\textcolor[rgb]{0,0.502,0}{\textsuperscript{-66}} & 16/313\textcolor[rgb]{0,0.502,0}{\textsuperscript{-41}} & 15/313\textcolor[rgb]{0,0.502,0}{\textsuperscript{-45}} & 112/313\textcolor[rgb]{0,0.502,0}{\textsuperscript{-45}} & 41/313\textcolor[rgb]{0,0.502,0}{\textsuperscript{-59}} & \textcolor[rgb]{0,0.502,0}{-52.5} \\
 & MPO & 16/313\textcolor[rgb]{0,0.502,0}{\textsuperscript{-45}} & 10/313\textcolor[rgb]{0,0.502,0}{\textsuperscript{-54}} & 10/313\textcolor[rgb]{0,0.502,0}{\textsuperscript{-54}} & 16/313\textcolor[rgb]{0,0.502,0}{\textsuperscript{-65}} & 20/313\textcolor[rgb]{0,0.502,0}{\textsuperscript{-37}} & 16/313\textcolor[rgb]{0,0.502,0}{\textsuperscript{-44}} & 67/313\textcolor[rgb]{0,0.502,0}{\textsuperscript{-90}} & 32/313\textcolor[rgb]{0,0.502,0}{\textsuperscript{-68}} & \textcolor[rgb]{0,0.502,0}{-62.66} \\
 & MPO+Our & 6/313\textcolor[rgb]{0,0.502,0}{\textsuperscript{-55}} & 5/313\textcolor[rgb]{0,0.502,0}{\textsuperscript{-59}} & 3/313\textcolor[rgb]{0,0.502,0}{\textsuperscript{-61}} & 4/313\textcolor[rgb]{0,0.502,0}{\textsuperscript{-77}} & 5/313\textcolor[rgb]{0,0.502,0}{\textsuperscript{-52}} & 5/313\textcolor[rgb]{0,0.502,0}{\textsuperscript{-55}} & 25/313\textcolor[rgb]{0,0.502,0}{\textsuperscript{-132}} & 7/313\textcolor[rgb]{0,0.502,0}{\textsuperscript{-93}} & \textcolor[rgb]{0,0.502,0}{-73.0} \\ 
\midrule
\multirow{4}{*}{Qwen2.5-7B} & {\cellcolor[rgb]{0.753,0.753,0.753}}None & {\cellcolor[rgb]{0.753,0.753,0.753}}16/313 & {\cellcolor[rgb]{0.753,0.753,0.753}}12/313 & {\cellcolor[rgb]{0.753,0.753,0.753}}21/313 & {\cellcolor[rgb]{0.753,0.753,0.753}}39/313 & {\cellcolor[rgb]{0.753,0.753,0.753}}27/313 & {\cellcolor[rgb]{0.753,0.753,0.753}}21/313 & {\cellcolor[rgb]{0.753,0.753,0.753}}98/313 & {\cellcolor[rgb]{0.753,0.753,0.753}}48/313 & {\cellcolor[rgb]{0.753,0.753,0.753}}- \\
 & Our & 3/313\textcolor[rgb]{0,0.502,0}{\textsuperscript{-13}} & 5/313\textcolor[rgb]{0,0.502,0}{\textsuperscript{-7}} & 6/313\textcolor[rgb]{0,0.502,0}{\textsuperscript{-15}} & 9/313\textcolor[rgb]{0,0.502,0}{\textsuperscript{-30}} & 12/313\textcolor[rgb]{0,0.502,0}{\textsuperscript{-15}} & 6/313\textcolor[rgb]{0,0.502,0}{\textsuperscript{-15}} & 60/313\textcolor[rgb]{0,0.502,0}{\textsuperscript{-38}} & 24/313\textcolor[rgb]{0,0.502,0}{\textsuperscript{-24}} & \textcolor[rgb]{0,0.502,0}{-19.625} \\
 & MPO & 6/313\textcolor[rgb]{0,0.502,0}{\textsuperscript{-10}} & 5/313\textcolor[rgb]{0,0.502,0}{\textsuperscript{-7}} & 5/313\textcolor[rgb]{0,0.502,0}{\textsuperscript{-16}} & 8/313\textcolor[rgb]{0,0.502,0}{\textsuperscript{-31}} & 11/313\textcolor[rgb]{0,0.502,0}{\textsuperscript{-16}} & 7/313\textcolor[rgb]{0,0.502,0}{\textsuperscript{-14}} & 25/313\textcolor[rgb]{0,0.502,0}{\textsuperscript{-73}} & 17/313\textcolor[rgb]{0,0.502,0}{\textsuperscript{-31}} & \textcolor[rgb]{0,0.502,0}{-28.0} \\
 & MPO+Our & 0/313\textcolor[rgb]{0,0.502,0}{\textsuperscript{-16}} & 0/313\textcolor[rgb]{0,0.502,0}{\textsuperscript{-12}} & 1/313\textcolor[rgb]{0,0.502,0}{\textsuperscript{-20}} & 2/313\textcolor[rgb]{0,0.502,0}{\textsuperscript{-37}} & 3/313\textcolor[rgb]{0,0.502,0}{\textsuperscript{-24}} & 1/313\textcolor[rgb]{0,0.502,0}{\textsuperscript{-20}} & 11/313\textcolor[rgb]{0,0.502,0}{\textsuperscript{-87}} & 11/313\textcolor[rgb]{0,0.502,0}{\textsuperscript{-37}} & \textcolor[rgb]{0,0.502,0}{-31.625} \\
\bottomrule
\end{tabular}
}
\end{sc}
\end{small}
\end{table}

Table~\ref{tab:main_multilingual_results_full} provides the complete safety results for all eight languages in \textsc{Multi-StrongREJECT}, complementing the subset reported in Table~\ref{tab:main_multilingual_results} in the main text.
Consistent with our main findings, our training-free update reduces unsafe completions across most languages and backbones, and composes well with MPO (often yielding the lowest unsafe-response counts).

\end{document}